%% file: GreedyAlg.tex
\documentclass{article}
\usepackage{graphicx} 
 \usepackage{tikz}
 \usepackage{booktabs}
 \usepackage{natbib}
 \usepackage{pgfplots}
\usetikzlibrary{3d}
\usetikzlibrary{backgrounds}
\pgfdeclarelayer{bg}
\pgfsetlayers{bg,main}
\usepackage[left=1.5cm, right=2cm]{geometry}

 \newgeometry{left=3cm,bottom=4cm}

\tikzset{
  background/.style={%
    execute at begin node={\begin{pgfonlayer}{bg}},
    execute at end node={\end{pgfonlayer}}
  }
}

  \usepackage{algorithm,algpseudocode}
  \usepackage{tikz}
  \usetikzlibrary{calc}
  \usepackage{tkz-euclide}
  \usepackage{xcolor}
\usetikzlibrary{arrows}
\usetikzlibrary{shapes.geometric, arrows.meta}
\usepackage{amsmath}
  
 \usetikzlibrary{math}
\title{$O(\sqrt{T})$ Static Regret and Instance Dependent Constraint Violation for Constrained Online Convex Optimization}
\author{%
  Rahul Vaze, Abhishek Sinha,  \\
 School of Technology and Computer Science \\
  Tata Institute of Fundamental Research \\
  Mumbai 400005, India \\
  \texttt{rahul.vaze@gmail.com},
  \texttt{abhishek.sinha@tifr.res.in}
}
\input{input.tex}

\begin{document}

\maketitle
\begin{abstract} The constrained version of the standard online convex optimization (OCO) framework, called COCO is considered, where on every round, a convex cost function and a convex constraint function are revealed to the learner after it chooses the action for that round.
The objective is to simultaneously minimize the static regret and cumulative constraint violation (CCV). 
An algorithm is proposed that guarantees a static regret of $O(\sqrt{T})$ and a CCV of $\min\{\cV, O(\sqrt{T}\log T) \}$, where $\cV$ depends on the distance between the consecutively revealed constraint sets, the shape of constraint sets, dimension of action space and the diameter of the action space. For special cases of constraint sets, $\cV=O(1)$. Compared to the state of the art results, static regret of $O(\sqrt{T})$ and CCV of $O(\sqrt{T}\log T)$, that were universal, the new result on CCV is instance dependent, which is derived by exploiting the geometric properties of the constraint sets.
\end{abstract}
\input{Introduction}
\input{COCO}

\input{NeuRipsAlg}

\input{Algorithm}

\input{EquivalentConvexBodyProblem}
\input{recursiveUnsynched}

\input{Conclusions}
\bibliography{OCO.bib} 
\input{App-ProofRegret}
\input{App-ProofAvgWidth}
\input{App-Switch}
\input{app-2D}

 \end{document}

%% file: input.tex
\usepackage{amsfonts}
\usepackage{times}
\usepackage{latexsym}
\usepackage{amssymb}
\usepackage{amsmath}
\usepackage{verbatim}
\newtheorem{theorem}{Theorem}

\newtheorem{assumption}[theorem]{Assumption}

\newtheorem{rem}{Remark}
\newtheorem{definition}[theorem]{Definition}
\newtheorem{example}[theorem]{Example}

\newtheorem{lemma}[theorem]{Lemma}
\newenvironment{proof}{ \textbf{Proof:} }{ \hfill $\Box$}

\def\bb0{{\mathbb{0}}}

\def\bb{{\mathbf{b}}}

\def\b0{{\mathbf{0}}}

\def\b1{{\mathbf{1}}}

\def\bbR{{\mathbb{R}}}
\def\bbS{{\mathbb{S}}}

\def\cB{\mathcal{B}}
\def\cC{\mathcal{C}}

\def\cH{\mathcal{H}}

\def\cP{\mathcal{P}}

\def\cR{\mathcal{R}}
\def\cS{\mathcal{S}}
\def\cT{\mathcal{T}}
\def\cU{\mathcal{U}}
\def\cV{\mathcal{V}}

\def\cX{\mathcal{X}}

\def\sfT{\mathsf{T}}

\def\sfc{{\mathsf{c}}}

\def\sf0{{\mathsf{0}}}

\def\nn{\nonumber}

%% file: Introduction.tex
\section{Introduction}
In this paper, we consider the constrained version of the standard online convex optimization (OCO) framework, called constrained OCO or COCO. In COCO, on every round $t,$ the online algorithm first chooses an admissible action $x_t \in \mathcal{X} \subset \bbR^d,$ 
 and then the adversary chooses a convex loss/cost function $f_t: \mathcal{X} \to \mathbb{R}$ and a constraint function of the form $g_{t}(x) \leq 0,$ where $g_{t}: \mathcal{X} \to \mathbb{R}$ is a convex function. Since $g_{t}$'s are revealed after the action $x_t$ is chosen, an online algorithm need not necessarily take feasible actions on each round, and in addition to the static regret 
 \begin{eqnarray} \label{intro-regret-def}
	\textrm{Regret}_{[1:T]} \equiv \sup_{\{f_t\}_{t=1}^T} \sup_{x^\star \in \mathcal{X}} \textrm{Regret}_T(x^\star), ~\textrm{where~}\textrm{Regret}_T(x^\star) \equiv \sum_{t=1}^T f_t(x_t) - \sum_{t=1}^T f_t(x^\star),
\end{eqnarray}
an additional metric of interest  is the total cumulative constraint violation (CCV)  defined as 
  \begin{eqnarray} \label{intro-gen-oco-goal}
 	\textrm{CCV}_{[1:T]}  \equiv \sum_{t=1}^T \max(g_{t}(x_t),0). 
	\end{eqnarray}
Let $\mathcal{X}^\star$ be the feasible set consisting of all admissible actions that satisfy all constraints $g_{t}(x) \leq 0, t\in [T]$. Under the standard assumption that $\mathcal{X}^\star$ is not empty (called the {\it feasibility assumption}), the goal is to design an online algorithm to simultaneously achieve a small regret \eqref{intro-regret-def} with respect to any admissible benchmark $x^\star \in \mathcal{X}^\star$ and a small CCV \eqref{intro-gen-oco-goal}. 
%

With constraint sets 
${\cal G}_t = \{x\in \cX : g_t(x)\le 0\}$ being convex for all $t$, and the assumption $\mathcal{X}^\star = \cap_t G_t  \neq \varnothing $  implies that sets
$S_t = \cap_{\tau=1}^t {\cal G}_\tau$ are convex and are nested, i.e. $S_t\subseteq S_{t-1}$ and $\mathcal{X}^\star \in S_t$ for all $t$. Essentially, set $S_t$'s are sufficient to quantify the CCV.

\subsection{Prior Work}
 {\bf Constrained OCO (COCO): (A) Time-invariant constraints:} COCO with time-invariant constraints, \emph{i.e.,} $g_{t} = g, \forall \ t$ \citep{yuan2018online, jenatton2016adaptive, mahdavi2012trading, yi2021regret} has been considered extensively, where functions $g$ are assumed to be known to the algorithm \emph{a priori}. The algorithm is allowed to take actions that are infeasible at any time to avoid the costly projection step of the vanilla projected OGD algorithm and the main objective was to design an \emph{efficient} algorithm  with a small regret and CCV while avoiding  the explicit projection step. 

{\bf (B) Time-varying constraints:} The more difficult question is solving COCO problem when the constraint functions, \emph{i.e.}, $g_{t}$'s, change arbitrarily with time $t$.  
In this setting, all prior work on COCO made the feasibility assumption.
One popular algorithm for solving COCO considered a Lagrangian function optimization that is updated using the primal and dual variables \citep{yu2017online, pmlr-v70-sun17a, yi2023distributed}. Alternatively, \citet{neely2017online} and \cite{georgios-cautious} used the drift-plus-penalty (DPP) framework  \citet{neely2010stochastic} to solve the COCO, but which needed additional assumption, e.g. the Slater's condition in \citet{neely2017online} and with weaker form of the feasibility assumption \cite{neely2017online}'s. 

More recently paper, \citet{guo2022online} obtained the bounds similar to \citet{neely2017online} but without assuming Slater's condition. However, the algorithm \citet{guo2022online} was quite computationally intensive since it requires solving a convex optimization problem on each round. 
Finally, very recently, the state of the art guarantees on simultaneous bounds on regret $O (\sqrt{T})$ and CCV $O (\sqrt{T}\log T)$ for COCO were derived in \cite{Sinha2024} with a very simple algorithm that combines the loss function at time $t$ and the CCV accrued till time $t$ in a single loss function, and then executes the online gradient descent (OGD) algorithm on the single loss function with an adaptive step-size.
Please refer to Table \ref{gen-oco-review-table} for a brief summary of the prior results.


The COCO problem has been considered in the {\it dynamic} setting as well  \citep{chen2018bandit, cao2018online, vazecocowiopt2022, liu2022simultaneously} where the benchmark $x^\star$ in \eqref{intro-regret-def} is replaced by $x_t^\star$ ($x_t^\star = \arg \min_x f_t(x)$) that is also allowed to change its actions over time. However, in this paper, we focus our entire attention on the static version.
A special case of COCO is the 
online constraint satisfaction (OCS) problem that does not involve any cost function, \emph{i.e.,} $f_t=0, \ \forall t,$ and the only object of interest is minimizing the CCV. The algorithm with state of the art guarantee for COCO \cite{Sinha2024} was shown to have a CCV of $O(\sqrt{T}\log T)$ for the OCS. 

\subsection{Convex Body Chasing Problem}  \label{cbc}
A well-studied problem related to the COCO is the
{\it nested convex body chasing (NCBC)} problem \citep{bansa2018nested,argue2019nearly,bubeck2020chasing}, 
where at each round $t$, a convex set $\chi_t \subseteq \chi$ is revealed such that 
$\chi_t\subseteq \chi_{t-1}$, and  $\chi_0=\chi \subseteq {\mathbb R}^d$ is a convex, compact, and bounded set. 
The objective is to choose action  $x_t \in \chi_t$ so as to minimize the total movement cost 
$C =   \sum_{t=1}^T  ||x_t - x_{t-1}||,$
where $x_0 \in \chi$ is some fixed action. Best known-algorithms for NCBC \citep{bansa2018nested,argue2019nearly,bubeck2020chasing} choose $x_t$ to be the 
centroid or Stiener point of $\chi_t$, essentially well inside the newly revealed convex set in order to reduce the future movement cost. 
With COCO, such an approach does not provide any useful bounds because of the presence of cost functions $f_t$'s whose minima could be towards the boundary of relevant convex sets $S_t$'s.

\subsection{Limitations of Prior Work}
We explicitly show in Lemma \ref{lem:algwc} that the best known algorithm \cite{Sinha2024} for solving COCO suffers a CCV of $\Omega(\sqrt{T}\log T)$ for `simple' problem instances where $f_t=f$ and $g_t=g$ for all $t$ and $d=1$ dimension, for which ideally the CCV should be $O(1)$. The same is true for most other prior algorithms, where the main reason for their large CCV for simple instances is that all these algorithms treat minimizing the CCV as a regret minimization problem for functions $g_t$. What they fail to exploit is the geometry of the 
underlying nested convex sets $S_t$'s  that control the CCV.
\subsection{Main open question}

In comparison to the above discussed upper bounds, the best known simultaneous lower bound \cite{Sinha2024}  for COCO is $\cR_{[1:T]} = \Omega(\sqrt{d})$ and $\text{CCV}_{[1:T]} = \Omega(\sqrt{d})$, where $d$ is the dimension of the action space $\cX$. Without constraints, i.e., $g_t\equiv0$ for all $t$, the lower bound on $\cR_{[1:T]} = \Omega(\sqrt{T})$ \cite{Hazan}.
Thus, there is a fundamental gap between the lower and upper bound for the CCV, and the main open question for COCO is : 

{\it  Is it possible to simultaneously achieve $\cR_{[1:T]} =O(\sqrt{T})$ and $\text{CCV}_{[1:T]} = o(\sqrt{T})$ or $\text{CCV}_{[1:T]} = O(1)$} for COCO?

Even though we do not fully resolve this question, in this paper, we make some meaningful progress by proposing an algorithm that exploits the geometry of the nested sets $S_t$'s and show that it is possible to simultaneously achieve  $\cR_{[1:T]} =O(\sqrt{T})$ and $\text{CCV}_{[1:T]} = O(1)$ in certain cases, and for general case, give a bound on the CCV that depends on the shape of the convex sets $S_t$'s while achieving  $\cR_{[1:T]} =O(\sqrt{T})$. In particular, the contributions of this paper are as follows.

\subsection{Our Contributions}
In this paper, we propose an algorithm (Algorithm \ref{coco_alg_1}) that tries to exploit the geometry of the nested convex sets $S_t$'s. In particular, Algorithm \ref{coco_alg_1} at time $t$, first takes an OGD step from the previous action $x_{t-1}$ with respect to the most recently revealed loss function $f_{t-1}$ with appropriate step-size to reach $y_{t-1}$, and then projects $y_{t-1}$ onto  the most recently revealed set $S_{t-1}$ to get $x_t$,  the action to be played at time $t$. 
Let $F_t$ be the ``projection" hyperplane passing through $x_t$ that is perpendicular to $x_t-y_{t-1}$. For Algorithm \ref{coco_alg_1}, we derive the following guarantees.
\begin{itemize}
\item The regret of the Algorithm \ref{coco_alg_1} is $O(\sqrt{T})$.
\item The CCV for the Algorithm \ref{coco_alg_1} takes the following form 
\begin{itemize}
\item When sets $S_t$'s are `nice', e.g. are spheres, or axis parallel cuboids, CCV is $O(1)$.
\item For general $S_t$'s, the CCV is upper bounded by a quantity $\cV$ that is a function of the distance between the consecutive sets $S_t$ and $S_{t+1}$ for all $t$, the shape of $S_t$'s, dimension $d$ and the diameter $D$.  Since $\cV$ depends on the shape of $S_t$'s, there is no universal bound on $\cV$, and the derived bound is instance dependent.
\item 
For the special case of $d=2$, when projection hyperplanes $F_t$'s progressively make increasing angles with respect to the first projection hyperplane $F_1$, the CCV is $O(1)$.
\end{itemize}
\item As pointed out above, for general $S_t$'s, there is no universal bound on the CCV of Algorithm \ref{coco_alg_1}. 
Thus, we propose an algorithm $\mathrm{Switch}$  that combines Algorithm \ref{coco_alg_1} and the algorithm from \cite{Sinha2024} to provide a regret bound of $O(\sqrt{T})$ and a CCV that is minimum of $\cV$ and $O(\sqrt{T} \log T)$. Thus, $\mathrm{Switch}$ provides a best of two worlds  CCV guarantee, which is small if the sets $S_t$'s are nice, while in the worst case it is at most $O(\sqrt{T} \log T)$.
\item For the OCS problem, we show that the CCV of Algorithm \ref{coco_alg_1} is $O(1)$ compared to the CCV of $O(\sqrt{T} \log T)$ \cite{Sinha2024}.
\end{itemize}

 \begin{table*}[t]
  \begin{tabular}{llllll}
    \toprule
   \small { Reference}  & \small {Regret} & \small {CCV} & \small {Complexity per round}\\
    \midrule
    \small {\citet{neely2017online}}  & \small {$O(\sqrt{T})$} & \small {$O(\sqrt{T})$} & \small {Conv-OPT, \small {Slater's condition}} \\
    \small {\citet{guo2022online}}  & \small {$O(\sqrt{T})$} & \small {$O(T^{\frac{3}{4}})$} & \small {Conv-OPT} \\
  \small {\citet{yi2023distributed}}  & \small {$O(T^{\max(\beta, 1-\beta)})$} & \small {$O(T^{1-\beta/2})$} & \small {Conv-OPT}  \\
     \small {\citet{Sinha2024}} & \small {$O(\sqrt{T})$} & \small {$O(\sqrt{T}\log T)$} & \small {Projection} \\
 \small {\textbf{This paper}} &   \small {$O(\sqrt{T})$} & \small {$O(\min\{\cV,\sqrt{T}\log T\})$} &\small {Projection}  \\
       \bottomrule
  \end{tabular}
  \vspace{5pt}
  \caption{\small{Summary of the results on COCO for arbitrary time-varying convex constraints and convex cost functions. In the above table, $0\leq \beta \leq 1$ is an adjustable parameter. Conv-OPT refers to solving a constrained convex optimization problem on each round. Projection refers to the Euclidean projection operation on the convex set $\mathcal{X}$. The CCV bound for this paper is stated in terms of $\cV$ which can be $O(1)$ or depends on the shape of convex sets $S_t$'s. }}
    \label{gen-oco-review-table}
\end{table*}

%% file: COCO.tex
\section{COCO Problem}

On round $t,$ the online policy first chooses an admissible action $x_t \in \mathcal{X}\subset \bbR^d,$ 
 and then the adversary chooses a convex cost function $f_t: \mathcal{X} \to \mathbb{R}$ and a constraint of the form $g_{t}(x) \leq 0,$ where $g_{t}: \mathcal{X} \to \mathbb{R}$ is a convex function. Once the action $x_t$ has been chosen, we let $\nabla f_t(x_t)$ and full function $g_t$ or the set $\{x: g_t(x)\le 0\}$ to be revealed, as is standard in the literature.
 We now state the  standard assumptions made  in the  literature while studying the COCO problem \cite{guo2022online, yi2021regret, neely2017online, Sinha2024}.
\begin{assumption}[Convexity] \label{cvx}
$\mathcal{X} \subset \bbR^d$ is the admissible set that is closed, convex and has a finite Euclidean diameter $D$.  The cost function $f_t: \mathcal{X} \mapsto \mathbb{R}$ and the constraint function $g_{t}: \mathcal{X} \mapsto \mathbb{R}$ are convex for all $t\geq 1$.  
\end{assumption}
\begin{assumption}[Lipschitzness] \label{bddness}
All cost functions $\{f_t\}_{t\geq 1}$ and the constraint functions $\{g_{t}\}_{ t\geq 1}$'s are $G$-Lipschitz, i.e., for any $x, y \in \mathcal{X},$ we have 
 \begin{eqnarray*}
 	|f_t(x)-f_t(y)| \leq G||x-y||,~
 	|g_{t}(x)-g_{t}(y)| \leq G||x-y||, ~\forall t\geq 1.
 \end{eqnarray*}
	\end{assumption}
 \begin{assumption}[Feasibility] \label{feas-constr}
With ${\cal G}_t = \{x\in \cX : g_t(x)\le 0\}$, we assume that  $\mathcal{X}^\star = \cap_{t=1}^T G_t  \neq \varnothing $.	
Any action $x^\star \in \cX^\star$ is defined to be feasible. 
\end{assumption}
The feasibility assumption distinguishes the cost functions from the constraint functions and is common across all previous literature on COCO \cite{guo2022online, neely2017online, yu2016low,yuan2018online,yi2023distributed, georgios-cautious,Sinha2024}. 


For any real number $z$, we define $(z)^+ \equiv \max(0,z).$ Since $g_{t}$'s are revealed after the action $x_t$ is chosen, any online policy need not necessarily take feasible actions on each round. 
 Thus in addition to the static\footnote{ The static-ness refers to the fixed benchmark using only one action $x^\star$ throughout the horizon of length $T$}  regret defined below
\begin{eqnarray} \label{regret-def}
	\textrm{Regret}_{[1:T]} \equiv \sup_{\{f_t\}_{t=1}^T} \sup_{x^\star \in \mathcal{X}^\star} \textrm{Regret}_{[1:T]}(x^\star), \end{eqnarray}
	where $\textrm{Regret}_{[1:T]}(x^\star) \equiv \sum_{t=1}^T f_t(x_t) - \sum_{t=1}^T f_t(x^\star)$, 
an additional obvious metric of interest is  the total cumulative constraint violation (CCV) defined as 
  \begin{eqnarray} \label{gen-oco-goal}
 	\textrm{CCV}_{[1:T]}  = \sum_{t=1}^T (g_{t}(x_t))^+. 
	\end{eqnarray}
	 Under the standard assumption (Assumption \ref{feas-constr}) that $\mathcal{X}^\star$ is not empty, the goal is to design an online policy to simultaneously achieve a small regret \eqref{intro-regret-def} with $x^\star \in \mathcal{X}^\star$ and a small CCV \eqref{intro-gen-oco-goal}. We refer to this problem as the constrained OCO (COCO). 
	 
For simplicity, we define set 
\begin{equation}\label{defn:S}
 \ S_t = \cap_{\tau=1}^t {\cal G}_\tau,
\end{equation}
where $G_t$ is as defined in Assumption \ref{feas-constr}.
All ${\cal G}_t$'s are convex and consequently, all $S_t$'s are convex and are nested, i.e. $S_t\subseteq S_{t-1}$. Moreover, because of Assumption \ref{feas-constr},  each $S_t$ is non-empty and in particular $\cX^\star\in S_t$ for all $t$. After action $x_t$ has been chosen, set $S_t$ controls the constraint violation, which can be used to write an upper bound on the $\textrm{CCV}_{[1:T]}$ as follows.

\begin{definition}
For a convex set $\chi$ and a point $x\notin \chi$, 
$$\text{dist}(x,\chi) = \min_{y\in \chi} || x-y||.$$
\end{definition}

Thus, the constraint violation at time $t$, 
\begin{equation}\label{eq:distviolationrelation}
(g_t(x_t))^+ \le G\text{dist}(x_t,S_t), \ \text{and} \  \textrm{CCV}_{[1:T]}  \le G\sum_{t=1}^T \text{dist}(x_t,S_t),
\end{equation}
where $G$ is the common Lipschitz constants for all $g_t$'s.

%% file: NeuRipsAlg.tex
\section{Algorithm from \cite{Sinha2024}}
The best known algorithm (Algorithm \ref{coco_sinha}) to solve COCO \cite{Sinha2024} was shown to have the following guarantee. 
\begin{theorem}\label{thm:sinha2024}[\cite{Sinha2024}]
Algorithm \ref{coco_sinha}'s $\textrm{Regret}_{[1:T]} = O(\sqrt{T})$ and  $\textrm{CCV}_{[1:T]} = O(\sqrt{T}\log T)$ when $f_t,g_t$ are convex.
\end{theorem}
We next show that in fact the analysis of \cite{Sinha2024} is tight for the CCV even when $d=1$ and $f_t(x)=f(x)$ and $g_t(x) =g(x)$ for all $t$. 
With finite diameter $D$ and the fact that any $x^\star \in \cX^\star$ belongs to all nested convex bodies $S_t$'s, when $d=1$, one expects that the CCV for any algorithm in this case will be $O(D)$. However, we as we show next,  Algorithm \ref{coco_sinha} does not effectively make use of geometric constraints imposed by nested convex bodies $S_t$'s.

\begin{algorithm}[tb]
   \caption{Online Algorithm from  \cite{Sinha2024}}
   \label{coco_sinha}
\begin{algorithmic}[1]
   \State {\bfseries Input:} Sequence of convex cost functions $\{f_t\}_{t=1}^T$ and constraint functions $\{g_t\}_{t=1}^T,$ $G=$ a common Lipschitz constant, $T=$ Horizon length,
    $D=$ Euclidean diameter of the admissible set $\mathcal{X},$ $\mathcal{P}_\mathcal{X}(\cdot)=$ Euclidean projection oracle on the set $\mathcal{X}$ 
     \State {\bfseries Parameter settings:} 
     \begin{enumerate}
     	\item \textbf{Convex cost functions:} $\beta = (2GD)^{-1}, V=1, \lambda = \frac{1}{2\sqrt{T}}, \Phi(x)= \exp(\lambda x)-1.$
     
    \item \textbf{$\alpha$-Strongly convex cost functions:} $\beta =1, V=\frac{8G^2 \ln(Te)}{\alpha}, \Phi(x)= x^2.$
    \end{enumerate}
  \State {\bfseries Initialization:} Set $ x_1={\bf 0}, \text{CCV}(0)=0$.
   \State {\bf For} $t=1:T$
   \State \quad Play $x_t,$ observe $f_t, g_t,$ incur a cost of $f_t(x_t)$ and constraint violation of $(g_t(x_t))^+$
   \State \quad $\tilde{f}_t \gets \beta f_t, \tilde{g}_t \gets \beta \max(0,g_t).$
   \State \quad $\text{CCV}(t)=\text{CCV}(t-1)+\tilde{g}_t(x_t).$
   \State \quad Compute $\nabla_t = \nabla \hat{f}_t(x_t),$ where $\hat{f}_t(x):= V\tilde{f}_t(x)+ \Phi'(\text{CCV}(t)) \tilde{g}_t(x), ~~ t \geq 1$.
   \State \quad $x_{t+1} = \mathcal{P}_\mathcal{X}(x_t - \eta_t \nabla_t)$, where 
   \quad \begin{eqnarray*}
   \eta_t =\begin{cases}
   	\frac{\sqrt{2}D}{2\sqrt{\sum_{\tau=1}^{t} ||\nabla_\tau||_2^2}}, ~&~\textrm{for convex costs} \\
   	\frac{1}{\sum_{s=1}^t H_s}, ~ &~ \textrm{for strongly convex costs } (H_t \textrm{ is the strong convexity parameter of } f_t). 
   	\end{cases}
   	\end{eqnarray*}
   	
   \State {\bf EndFor}
\end{algorithmic}
\end{algorithm}

\begin{lemma}\label{lem:algwc}
Even when $d=1$ and $f_t(x)=f(x)$ and $g_t(x) =g(x)$ for all $t$, for Algorithm \ref{coco_sinha}, its $\textrm{CCV}_{[1:T]}  = \Omega(\sqrt{T} \log T)$.
\end{lemma}
\begin{proof}
{\bf Input:} Consider $d=1$, and let $\cX=[1, a], a>2$. Moreover, let $f_t(x)=f(x)$ and $g_t(x) =g(x)$ for all $t$. Let $f(x) = c x^2$ for some (large) $c>0$ and $g(x)$ be such that $G=\{x: g(x)\le 0\} \subseteq [a/2, a]$ and let $|\nabla g(x)|\le1$ for all $x$.

Let $1< x_1 < a/2$. Note that $\text{CCV}(t)$ (defined in Algorithm \ref{coco_sinha}) is a non-decreasing function, and let $t^\star$ be the earliest time $t$ such that $\Phi'(\text{CCV}(t)) \nabla g(x) <- c $. 
For $f(x) = c x^2$, $\nabla f(x) \ge c$ for all $x>1$. 
Thus, using Algorithm \ref{coco_sinha}'s definition, it follows that for all $t\le t^\star$, $x_t < a/2$, since the derivative of $f$ dominates the derivative of $\Phi'(\text{CCV}(t))  g(x)$ until then.

Since  $\Phi(x)= \exp(\lambda x)-1$ with $\lambda = \frac{1}{2\sqrt{T}}$, and by definition $|\nabla g(x)| \le 1$ for all $x$, thus, we have that by time $t^\star$, 
$\textrm{CCV}_{[1:t^\star]} = \Omega(\sqrt{T} \log T)$. Therefore, $\textrm{CCV}_{[1:T]} =\Omega(\sqrt{T} \log T)$.

 \end{proof}
 
Essentially, Algorithm \ref{coco_sinha} is treating minimizing the $\text{CCV}$ problem as regret minimization for function $g$ similar to function $f$ and this leads to its CCV of $\Omega(\sqrt{T}\log T)$.
For any given input instance with $d=1$, an alternate algorithm that  chooses its actions following online gradient descent (OGD) projected on to the most recently revealed feasible set $S_t$ achieves $O(\sqrt{T})$ regret (irrespective of the starting action $x_1$) and $O(D)$ $\text{CCV}$ (since any $x^\star \in S_t$ for all $t$).  We extend this intuition in the next section, and present an algorithm that tries to exploit the geometry of the nested convex sets $S_t$ for general $d$.

%% file: Algorithm.tex
\section{New Algorithm for solving COCO}
In this section, we present a simple algorithm (Algorithm \ref{coco_alg_1}) for solving COCO.
\begin{algorithm}[tb]
   \caption{Online Algorithm for COCO}
   \label{coco_alg_1}
\begin{algorithmic}[1]
   \State {\bfseries Input:} Sequence of convex cost functions $\{f_t\}_{t=1}^T$ and constraint functions $\{g_t\}_{t=1}^T,$ $G=$ a common Lipschitz constant,  $d$ dimension  of the admissible set $\mathcal{X},$ step size $\eta_t = \frac{D}{G \sqrt{t}}$. 
    $D=$ Euclidean diameter of the admissible set $\mathcal{X},$ $\mathcal{P}_\mathcal{X}(\cdot)=$ Euclidean projection operator on the set $\mathcal{X}$,      \State {\bfseries Initialization:} Set $ x_1 \in \mathcal{X}$ arbitrarily, $\text{CCV}(0)=0$.
   \State {\bf For} \ {$t=1:T$}
   \State \quad Play $x_t,$ observe $f_t, g_t,$ incur a cost of $f_t(x_t)$ and constraint violation of $(g_t(x_t))^+$
   \State \quad Set $S_t$ as defined in \eqref{defn:S}
    \State \quad $y_{t} =  \mathcal{P}_{S_{t-1}}\left(x_t - \eta_t \nabla f_t(x_t)\right)$
   \State \quad $x_{t+1} =  \mathcal{P}_{S_t}\left(y_t\right)$
   \State {\bf EndFor}
\end{algorithmic}
\end{algorithm}
Algorithm \ref{coco_alg_1} is essentially an online projected gradient algorithm (OGD), 
which first takes an OGD step from the previous action $x_{t-1}$ with respect to the most recently revealed loss function $f_{t-1}$ with appropriate step-size which is then projected onto $S_{t-2}$ to reach $y_{t-1}$, and then projects $y_{t-1}$ onto  the most recently revealed set $S_{t-1}$ to get $x_t$,  the action to be played at time $t$.
\eqref{defn:S}. 

\begin{rem} Step 6 of Algorithm \ref{coco_alg_1} might appear unnecessary, however, its useful for proving Theorem \ref{thm:tvmonotone}.
\end{rem}

Since Algorithm \ref{coco_alg_1} is essentially an online projected gradient algorithm, similar to classical result on OGD, next, we show that the regret of Algorithm \ref{coco_alg_1} is $O(\sqrt{T})$.
\begin{lemma}\label{lem:regretbound}
The $\textrm{Regret}_{[1:T]}$ for Algorithm \ref{coco_alg_1} is $O(\sqrt{T})$.
\end{lemma}
Extension of Lemma \ref{lem:regretbound} when $f_t$'s are strongly convex which results in $\textrm{Regret}_{[1:T]}=O(\log{T})$ for Algorithm \ref{coco_alg_1} follows standard arguments \cite{Hazan} and is omitted.

The real challenge is to bound the total $\text{CCV}$ for Algorithm \ref{coco_alg_1}. 
Let $x_t$ be the action played by Algorithm \ref{coco_alg_1}. Then by definition, $x_t \in S_{t-1}$. Moreover, from \eqref{eq:distviolationrelation}, the constraint violation at time $t$, $\text{CCV}(t) \le G \text{dist}(x_{t}, S_t)$.
The next action $x_{t+1}$ chosen by Algorithm \ref{coco_alg_1} belongs to $S_t$, however, it is obtained by first taking an OGD step from $x_t$ to reach $y_t$ and then projects $y_t$ onto $S_t$. Since $f_t$'s are arbitrary, the OGD step could be towards any direction, and thus, there is no direct relationship between $x_{t+1}$ and $x_t$. Informally, $(x_1, x_2, \dots, x_T)$ is not a connected curve with any useful property. Thus, we take recourse in upper bounding the CCV via upper bounding the total movement cost $M$ (defined below) between nested convex sets using projections.

  The total constraint violation for Algorithm \ref{coco_alg_1} is
\begin{align}\nn
\text{CCV}_{[1:t]} & \le G\sum_{\tau=1}^t \text{dist}(x_{\tau}, S_{\tau}), \\ \label{defn:genconvxmovement}
&\stackrel{(a)} \le G  \sum_{\tau=1}^t ||x_{\tau}-  b_\tau||, \\
&\stackrel{(b)} = G M_t,
\end{align}
where in $(a)$ $b_t$ is the projection of $x_t$ onto $S_{t}$, i.e., $b_t=\cP_{S_{t}}(x_t)$ and in $(b)$
\begin{equation} \label{defn:totalmovementcost1}
M_t= \sum_{\tau=1}^t ||x_{\tau}-  b_\tau||
\end{equation} is defined to be the  total movement cost  on the instance $S_1, \dots, S_t$. 
The object of interest is $M_T$.

%% file: EquivalentConvexBodyProblem.tex
\section{Bounding the Total Movement Cost $M_T$ \eqref{defn:totalmovementcost1}}


We start by considering two simple cases where bounding $M_T$ is easy.
 \begin{lemma}\label{lem:spheres}
If all nested convex bodies $S_1\supseteq S_2  \supseteq \dots \supseteq S_T$ are spheres then $M_T\le d^{3/2}D$.
\end{lemma}

\begin{proof}
Recall the definition that $x_t\in \partial S_{t-1}, b_t=\cP_{S_{t}}(x_t)\in S_t$ from \eqref{defn:genconvxmovement}.
Let $||x_t-b_t||=r$, then since all $S_t$'s are spheres, at least along one of the $d$-orthogonal canonical basis vectors, $\text{diameter}(S_{t})\le \text{diameter}(S_{t-1}) - \frac{r}{\sqrt{d}}$. Since the diameter along any of the $d$-axis is $D$, we get the answer.
\end{proof}

 \begin{lemma}\label{lem:square}
If all nested convex bodies $S_1\supseteq S_2  \supseteq \dots \supseteq S_T$ are cuboids that are axis parallel to each other, then $M\le d^{3/2}D$.
\end{lemma}
Proof is identical to Lemma \ref{lem:spheres}.
Note that similar results can be obtained when $S_t$'s are regular polygons that are axis parallel with each other.

 After exhausting the universal results for an upper bound on $M_T$ for `nice' nested convex bodies, we next give a general bound on $M_T$ for any sequence of nested convex bodies which depends on the geometry of the nested convex bodies (instance dependent).
 To state the result we need the following preliminaries.
 
\begin{figure*}
\begin{center}
\includegraphics[width=10cm,keepaspectratio,angle=0]{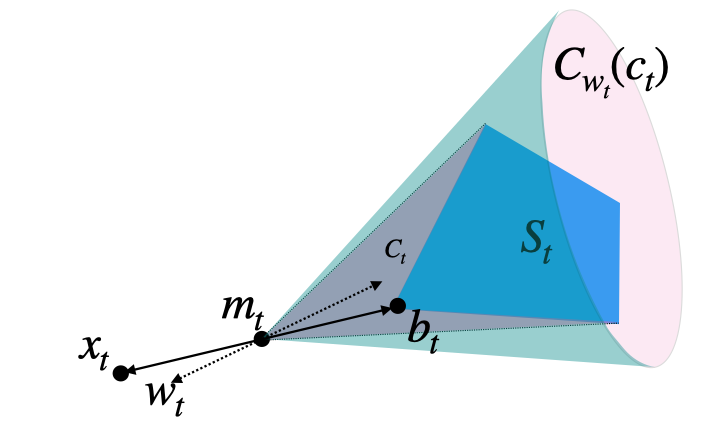}
\caption{Figure representing the cone $C_{w_t}(c_t)$ that contains the convex hull of $m_t$ and $S_{t}$ with unit vector $w_t$.} 
\label{fig:anglewidthmain}
\end{center}
\end{figure*}

 Following \eqref{defn:genconvxmovement}, $b_t=\cP_{S_t}(x_t)$ where $x_t\in \partial S_{t-1}$. Without loss of generality, 
 $x_t\notin S_{t}$ since otherwise the distance $||x_t-b_t||=0$.
 Let $m_t$ be the mid-point of $x_t$ and $b_t$, i.e. $m_t = \frac{x_t+b_t}{2}$.
 \begin{definition}\label{defn:anglewidth}
Let the convex hull of $m_t \cup S_{t}$ be $\cC_t$.
 Let $w_t$ be a unit vector such that there exists $c_t>0$ such that the cone 
 $$C_{w_t}(c_t) = \left\{z\in \bbR^d: -w_t^T\frac{(z-m_t)}{||(z-m_t)||} \ge c_t\right\}$$ contains $\cC_t$. Since $S_{t}$ is convex, such $w_t, c_t>0$ exist. For example, $w_t=b_t-x_t$ is one such choice for which $c_t>0$ since $m_t \notin S_t$.
See Fig. \ref{fig:anglewidthmain} for a pictorial representation.
 
 Let $c^\star_{w_t,t} = \arg \min_{c_t} C_{w_t}(c_t)$,
 $c^\star_t = \min_{w_t} c^\star_{w_t,t}$, and $w_t^\star= \arg \min_{w_t} c^\star_{w_t,t}$.
 Moreover, let $c^\star = \min_t  c^\star_t$, where by definition, $c^\star <1$. 
 \end{definition}

Essentially, $2\cos^{-1}(c^\star_t)$ is the angle width of $\cC_t$ with respect to $w_t^\star$, i.e. each element of $\cC_t$ makes an  angle of at most $ \cos^{-1}(c^\star_t)$ with $w_t^\star$.

 

\begin{rem}\label{rem:cbound}
Note that $c_t^\star$ is only a function of the distance $ ||x_t-b_t||$   and the shape of $S_t$'s, in particular, the maximum width of $S_t$ 
 along the directions perpendicular to vector $x_t-b_t$ $\forall \ t$ which can be at most the diameter $D$. 
 $c_t^\star$ decreases (increasing the ``width" of cone $C_{w_t^\star}(c_t^\star)$) as $||x_t-b_t||$ decreases, but small $x_t-b_t$ also implies small  violation at time $t$ from \eqref{defn:genconvxmovement}.
Across time slots, $d_{\min} = \min_t ||x_t-b_t||$ and shape of $S_t$'s control $c^\star$, where $d_{\min} > 0$ is inherent from the definition of $c^\star$ since a bound on $||x_t-b_t||$ is only needed for the case when $x_t\ne b_t$. 
\end{rem} 

\begin{rem}  Projecting $x_t\in \partial S_{t-1}$ onto $S_t$ to get $b_t=\cP_{S_t}(x_t)$, the diameter of $S_t$ is at most diameter of $S_{t-1} - ||x_t-b_t||$, however, only along the direction $b_t-x_t$. Since the shape of $S_{t}$ is arbitrary, as a result, the diameter of $S_t$ need not be smaller than the diameter of $S_{t-1}$ along any pre-specified direction, which was the main idea used to derive Lemma \ref{lem:spheres}.  Thus, to relate the distance $||x_t-b_t||$ with the decrease in the diameter of the convex bodies $S_t$'s, we use the concept of {\bf mean width} of a convex body that is defined as the expected width of the convex body along all the directions that are chosen uniformly randomly (formal definition is provided in Definition \ref{defn:avgwidth}).
\end{rem}

Next, we upper bound $M_T$ by connecting the distance $||x_t-b_t||$ to the decrease in mean width (to be defined ) of convex bodies $S_{t-1}$ and $S_t$'s.
 
 \begin{lemma}\label{lem:movementcost}
 The total movement cost 
$M_T$ in \eqref{defn:totalmovementcost1} is at most $$\frac{2V_d(d-1)}{V_{d-1}} \left(\frac{1}{c^\star}\right)^{d}D,$$ where 
$V_d$ is the $(d-1)$-dimensional Lebesgue measure of  the unit sphere in $d$ dimensions. 
\end{lemma}

Note that $V_d/V_{d-1} = O(1/\sqrt{d})$.
 Thus, we get the following {\bf main result} of the paper for Algorithm \ref{coco_alg_1} combining Lemma \ref{lem:regretbound} and Lemma \ref{lem:movementcost}. 
 \begin{theorem}\label{thm:main1}
For solving COCO, Algorithm \ref{coco_alg_1} has  $$\textrm{Regret}_{[1:T]} = O(\sqrt{T}), \ \text{and} \ \text{CCV}_{[1:T]}= O\left(\sqrt{d} \left(\frac{1}{c^\star}\right)^{d}D\right).$$
\end{theorem}

Compared to all prior results on COCO, that were universal (instance independent), where the best known one \cite{Sinha2024} has $\textrm{Regret}_{[1:T]} = O(\sqrt{T})$, and $\text{CCV}_{[1:T]}=O(\sqrt{T}\log T)$, Theorem \ref{thm:main1} is an instance 
dependent result for the CCV. In particular, it exploits the geometric structure of the nested convex sets $S_t$'s and 
derives an upper bound on the CCV that only depends on the `shape' of $S_t$'s. It can be the case that the instance is `badly' behaved and $c^\star$ is very small or dependent on $T$. If that is the case, in Section \ref{sec:algswitch} we show how to limit the CCV to $O(\sqrt{T}\log T)$. However, when $S_t$'s are `nice', e.g., $c^\star$ is independent of $T$ (Remark \ref{rem:cbound}) or $S_t$'s are spheres or axis parallel cuboids (Lemma \ref{lem:spheres} and \ref{lem:square}), the $\text{CCV}$ of Algorithm \ref{coco_alg_1} is independent of $T$, which is a fundamentally improved result compared to large body of prior work. In fact, in prior work this was largely assumed to be not possible. In particular, before the result of \cite{Sinha2024}, achieving simultaneous $\textrm{Regret}_{[1:T]} = O(\sqrt{T})$, and $\text{CCV}_{[1:T]}=O(\sqrt{T})$ itself was the final goal.

\section{Algorithm $\mathrm{Switch}$}\label{sec:algswitch}

Since Theorem \ref{thm:main1} provides an instance dependent bound on the CCV, that is a function of $c^\star$ which can be small, it can be the case that its CCV is larger than $O(\sqrt{T}\log T)$, thus providing a result that is inferior to that of Algorithm \ref{coco_sinha} \cite{Sinha2024}. Thus, next, we marry the two algorithms, Algorithm \ref{coco_sinha} and Algorithm \ref{coco_alg_1}, in Algorithm \ref{alg:switch} to provide a best of both results as follows.

\begin{algorithm}[tb]
   \caption{$\mathrm{Switch}$}
   \label{alg:switch}
\begin{algorithmic}[1]
   \State {\bfseries Input:} Sequence of convex cost functions $\{f_t\}_{t=1}^T$ and constraint functions $\{g_t\}_{t=1}^T,$ $G=$ a common Lipschitz constant,  $d$ dimension  of the admissible set $\mathcal{X},$
    $D=$ Euclidean diameter of the admissible set $\mathcal{X},$ $\mathcal{P}_\mathcal{X}(\cdot)=$ Euclidean projection operator on the set $\mathcal{X}$,      \State {\bfseries Initialization:} Set $ x_1 \in \mathcal{X}$ arbitrarily, $\text{CCV}(0)=0$.
   \State {\bf For} \ {$t=1:T$}
   \State \quad {\bf If} {$\text{CCV}(t-1) \le \sqrt{T}\log T$}
   \State \quad \quad Follow Algorithm \ref{coco_alg_1}
   \State  \quad  \quad $\text{CCV}(t)=\text{CCV}(t-1)+\max\{g_t(x_t),0\}.$
   \State \quad {\bf Else} 
   \State \quad \quad Follow Algorithm \ref{coco_sinha} with resetting $\text{CCV}(t-1)=0$
   \State \quad {\bf EndIf} 
   \State {\bf EndFor}
\end{algorithmic}
\end{algorithm}

 \begin{theorem}\label{thm:main2}
$\mathrm{Switch}$ (Algorithm \ref{alg:switch}) has regret $\textrm{Regret}_{[1:T]} =O(\sqrt{T})$, while    $$\text{CCV}_{[1:T]}=
\min\left\{O\left(\sqrt{d} \left(\frac{1}{c^\star}\right)^{d}D\right), O(\sqrt{T}\log T)\right\}.$$
\end{theorem}
 
Algorithm $\mathrm{Switch}$ should be understood as the best of two worlds algorithm, where the two worlds corresponds to one having nice convex sets $S_t$'s that have CCV independent of $T$ or $o(\sqrt{T})$ for 
Algorithm \ref{coco_alg_1}, while in the other, CCV of Algorithm \ref{coco_alg_1} is large on its own, and the overall CCV is controlled by discontinuing the use of Algorithm \ref{coco_alg_1} once its CCV reaches $\sqrt{T}\log T$ and switching to Algorithm \ref{coco_sinha} thereafter that has universal guarantee of $O(\sqrt{T}\log T)$ on its CCV.
 
 After exhausting the general results on the CCV of Algorithm \ref{coco_alg_1}, we next consider the special case of $d=2$ and when the sets $S_t$ have a special structure defined by their projection hyperplanes. 
 Note that it is highly non-trivial to bound the CCV of Algorithm \ref{coco_alg_1} even when $d=2$.


%% file: recursiveUnsynched.tex
\section{Special case of $d=2$}
In this section, we show that if $d=2$ (all convex sets $S_t$'s lie in a plane) and the projections satisfy a monotonicity property depending on the problem instance, then we can bound the total CCV for Algorithm \ref{coco_alg_1} independent of the time horizon $T$ and consequently getting a $O(1)$ CCV.

Recall from the definition of Algorithm \ref{coco_alg_1}, $y_t = \cP_{S_{t-1}}(x_t - \eta_t \nabla f_t(x_t))$ and 
$x_{t+1} = \cP_{S_t}(y_t)$.

\begin{figure}

\includegraphics[width=10cm,keepaspectratio,angle=0]{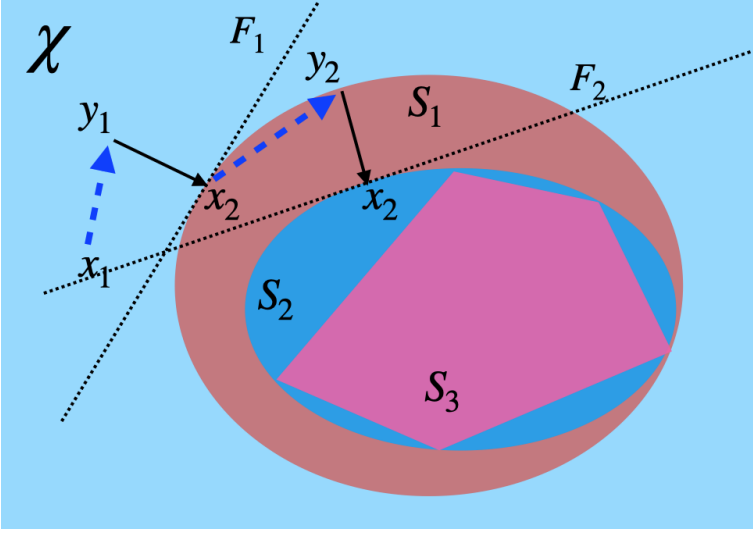}

\caption{Definition of $F_t$'s.}
\label{fig:defF}
\end{figure}

\begin{definition}\label{defn:projhyperplane}
Let the hyperplane perpendicular to line segment $(y_t, x_{t+1})$ passing through $x_{t+1}$ be 
$F_t$. Without loss of generality, we let $y_t \notin S_t$, since then the projection is trivial. Essentially $F_t$ is the projection hyperplane at time $t$. 
Let $\cH_t^+$ denote the positive half plane corresponding to $F_t$, i.e., 
$\cH_t^+ = \{z: z^T (y_t-x_{t+1})\ge 0\}$. 
Refer to Fig. \ref{fig:defF}.
Let the angle between $F_1$ and $F_t$ be $\theta_t$. 
\end{definition}

\begin{definition}\label{defn:anglemonotone}
The instance $S_1 \supseteq S_2 \supseteq \dots \supseteq S_T$ is defined to be monotonic 
if $\theta_2 \le \theta_3 \le \dots \le \theta_T$.
\end{definition}

\begin{theorem}\label{thm:tvmonotone}
For $d=2$ when the instance is monotonic, $\text{CCV}_{[1:T]}$ for Algorithm \ref{coco_alg_1} is at most $O(GD)$.
\end{theorem}

Theorem \ref{thm:tvmonotone} provides a universal guarantee on the CCV of  Algorithm \ref{coco_alg_1} that is independent of the problem instance (as long as it is monotonic) unlike Lemma \ref{lem:movementcost}, even though it applies only for $d=2$. The proof is derived by using basic convex geometry results from \cite{Manselli} in combination with exploiting the definition of Algorithm \ref{coco_alg_1} and the monotonicity condition. It is worth noting that even under the monotonicity assumption it is non-trivial to upper bound the CCV since the successive angles made by $F_t$ with $F_1$ can increase arbitrarily slowly, making it difficult 
to control the total CCV. 
\section{OCS Problem}
In \cite{Sinha2024}, a special case of COCO, called the OCS problem, was introduced where $f_t\equiv0$ for all $t$. Essentially, with OCS, only constraint satisfaction is the objective.  In \cite{Sinha2024}, Algorithm \ref{coco_sinha} was shown to have CCV of $O(\sqrt{T}\log T)$. Next, we show that Algorithm \ref{coco_alg_1} has CCV of $O(1)$ for the OCS, a remarkable improvement.

 \begin{theorem}\label{thm:ocs}
For solving OCS, Algorithm \ref{coco_alg_1} has $\text{CCV}_{[1:T]}= O\left(d^{d/2} D\right)$.
\end{theorem}

As discussed in \cite{Sinha2024}, there are important applications of OCS, and it is important to find tight bounds on its CCV. Theorem \ref{thm:ocs} achieves this by showing that CCV of $O(1)$ can be achieved, where the constant depends only on the dimension of the action space and the diameter. This is a fundamental improvement compared to the CCV bound of $O(\sqrt{T}\log T)$ from \cite{Sinha2024}. Theorem  \ref{thm:ocs} is derived by using the connection between the curve obtained by successive projections on nested convex sets and self-expanded curves (Definition \ref{defn:se-curve}) and then using a classical result on self-expanded curves from \cite{Manselli}.

%% file: Conclusions.tex
\section{Conclusions}
One fundamental open question for COCO is: whether it is possible to simultaneously achieve $\cR_{[1:T]} =O(\sqrt{T})$ and $\text{CCV}_{[1:T]} = o(\sqrt{T})$ or $\text{CCV}_{[1:T]} = O(1)$.
In this paper, we have made substantial progress towards answering this question by proposing an algorithm that exploits the geometric properties of the nested convex sets $S_t$'s that effectively 
control the CCV. The state of the art algorithm \cite{Sinha2024} achieves a CCV of $\Omega(\sqrt{T}\log T)$ even for very simple instances as shown in Lemma \ref{lem:algwc}, 
and conceptually different algorithms were needed to achieve CCV of $o(\sqrt{T})$. 
We propose one such algorithm and show that when the nested convex constraint sets are `nice' (instances  is simple), achieving a CCV of $O(1)$ is possible without losing out on $O(\sqrt{T})$ regret guarantee. We also derived a bound on the CCV for general problem instances, that is as a function of the shape of nested convex constraint sets and the distance between them, and the diameter. 

In the absence of good lower bounds, the open question remains open in general, however,
 this paper significantly improves the conceptual understanding of COCO problem by demonstrating that good algorithms need to exploit the geometry of the nested convex constraint sets.

One remark we want to make at the end is that COCO is inherently a difficult problem, which is best exemplified by the fact, that even for the special case of COCO where $f_t=f$ for all $t$, essentially where $f$ is known ahead of time, our algorithm/prior work does not yield a better regret or CCV bound compared to when $f_t$'s are arbitrarily varying. 

%% file: App-ProofRegret.tex
\newpage
\section{Proof of Lemma \ref{lem:regretbound}}
\begin{proof}
From the convexity of $f_t$'s, for $x^\star$ satisfying Assumption \eqref{feas-constr}, we have 
$$f_t(x_t) - f_t(x^\star) \le \nabla f_t^T (x_t-x^\star).$$
From the choice of Algorithm \ref{coco_alg_1} for $x_{t+1}$, we have 
\begin{align*} ||x_{t+1}-x^\star||^2 & = || \cP_{S_{t}}(y_t) - x^\star||^2 \\
& \stackrel{(a)}\le || y_t - x^\star||^2, \\
& = ||\cP_{S_{t-1}}\left(x_t - \eta_t \nabla f_t(x_t)\right)-x^\star||^2, \\
& \stackrel{(n)}\le || (x_t-\eta_t\nabla f_t^T(x_t)) - x^\star||^2,
\end{align*}
where inequalities $(a)$ and $(b)$ follow since $x^\star \in S_t$ for all $t$.
Hence
\begin{align*}
 ||x_{t+1}-x^\star||^2 & \le  ||x_t-x^\star||^2 + \eta_t^2||\nabla f_t(x_t)||^2 - 2\eta_t \nabla f_t^T(x_t)(x_t-x^\star), \\
 \nabla f_t^T(x_t)(x_t-x^\star) & \le \frac{||x_t-x^\star||^2-||x_{t+1}-x^\star||^2 }{\eta_t} + \eta_t G^2.
\end{align*}
Summing this over $t=1$ to $T$, we get 
\begin{align*}
2\sum_{t=1}^T (f_t(x_t) - f_t(x^\star)) & \le \sum_{t=1}^T\nabla f_t^T (x_t-x^\star), \\
& \le \sum_{t=1}^T  \frac{||x_t-x^\star||^2-||x_{t+1}-x^\star||^2 }{\eta_t} + \sum_{t=1}^T\eta_t G^2, \\
& \le D^2 \frac{1}{\eta_T} + G^2 \sum_{t=1}^T\eta_t,\\
& \le O( DG \sqrt{T}),
\end{align*}
where the final inequality follows by choosing $\eta_t = \frac{D}{G\sqrt{t}}$.
\end{proof}

%% file: App-ProofAvgWidth.tex
\section{Proof of Theorem \ref{lem:movementcost}}
\begin{proof}
We need the following preliminaries.

 \begin{definition}\label{defn:avgwidth}
Let $K$ be a non-empty convex bounded set in $\bbR^d$. Let $u$ be a unit vector, and $\ell_u$ a line through the origin parallel to $u$. 
Let $K_u$ be the orthogonal projection of $K$ onto $\ell_u$, with length $|K_u|$. The mean width of $K$ is defined as 
\begin{equation}\label{eq:projlength}
W(K) = \frac{1}{V_d} \int_{\bbS_1^d} |K_u| du,
\end{equation}
where $\bbS_1^d$ is the unit sphere in $d$ dimensions and $V_d$ its $(d-1)$-dimensional Lebesgue measure.
\end{definition}

The following is immediate. 
\begin{equation}\label{eq:WBound1}
0\le W(K) \le \text{diameter}(K).
\end{equation}

\begin{lemma}\label{lem:width2D}\cite{eggleston1966convexity}
For $d=2$, $$W(K)=\frac{\text{Perimeter}(K)}{\pi}.$$
\end{lemma}
Lemma \ref{lem:width2D} implies that $W(K) \ne W(K_1) + W(K_2)$ even if $K_1\cup K_2=K$ and $K_1\cap K_2=\phi$.

Recall from \eqref{defn:genconvxmovement} that $x_t\in  \partial S_{t-1}$ and $b_t$ is the projection of $x_t$ onto $S_{t}$, and $m_t$ is the mid-point of $x_t$ and $b_t$, i.e. $m_t = \frac{x_t+b_t}{2}$. Moreover, the convex sets $S_t$'s are nested, i.e., $S_1\supseteq S_2 \supseteq \dots \supseteq S_T$.
To prove Theorem \ref{lem:movementcost} we will bound the rate at which $W(S_t)$ (Definition \ref{defn:avgwidth}) decreases as a function of the length $||x_t-b_t||$. 

From Definition \ref{defn:anglewidth}, recall that $\cC_t$ is the convex hull of $m_t\cup S_{t}$. We also need to define $\cC_t^-$ as the convex hull of $x_t\cup S_{t}$. Since 
$S_t \subseteq \cC_t$ and $\cC_t^- \subseteq S_{t-1}$ (since $S_{t-1}$ is convex and $x_t\in S_{t-1}$), we have \begin{equation}\label{eq:widthlb}
W(S_{t}) - W(S_{t-1}) \le W(\cC_t) - W(\cC_t^-).
\end{equation} 
\begin{definition}\label{}
$\Delta_t = W(\cC_t) - W(\cC_t^-)$.
\end{definition}

The main ingredient of the proof is the following Lemma that bounds  $\Delta_t$ whose proof is provided after completing the proof of Theorem \ref{lem:movementcost}.
\begin{lemma}\label{lem:derivative} $$\Delta_t  \le -V_{d-1}\frac{||x_t-b_t||}{2V_d(d-1)}  (c_t^\star)^{d},$$
where $c_t^\star$ has been defined in Definition \ref{defn:anglewidth}.
\end{lemma}

Recalling that $c^\star = \min_t  c^\star_t$ from Definition \ref{defn:anglewidth}, and  combining  Lemma \ref{lem:derivative} with \eqref{eq:WBound1} and \eqref{eq:widthlb}, we get that 
$$\sum_{t=1}^T ||x_t-b_t|| \le \frac{2V_d(d-1)}{V_{d-1}} \left(\frac{1}{c^\star}\right)^{d}\text{diameter}(S_1),$$
since $S_1\supseteq S_2 \supseteq \dots \supseteq S_T$. Recalling that $\text{diameter}(S_1)\le D$, Theorem \ref{lem:movementcost} follows.
\end{proof}

\begin{proof}[Proof of Lemma \ref{lem:derivative}]

\begin{figure*}
\begin{center}
\includegraphics[width=10cm,keepaspectratio,angle=0]{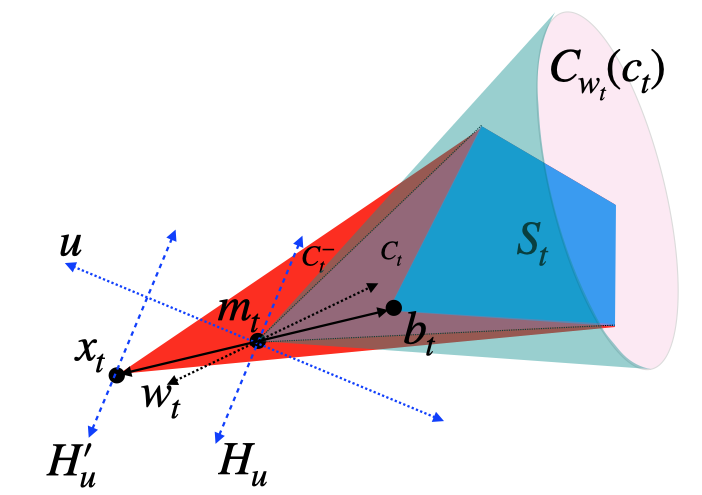}
\caption{Figure representing the cone $C_{w_t}(c_t)$ that contains the convex hull of $m_t$ and $S_{t}$ with respect to the unit vector $w_t$. $u$ is a unit vector perpendicular to $H_u$ an hyperplane that is a supporting hyperplane $C_t$ at $m_t$ such that $\cC_t \cap H_u = \{m_t\}$ and 
$u^T (x_t-m_t)\ge 0$ }
\label{fig:anglewidth}
\end{center}
\end{figure*}

Let $H_u$ be the hyperplane perpendicular to vector $u$.
 Let $\cU_0$ be the set of unit vectors $u$ such that hyperplanes $H_u$ are supporting hyperplanes to $\cC_t$ at point $m_t$ such that $\cC_t \cap H_u = \{m_t\}$ and 
$u^T (x_t-m_t)\ge 0$.  See Fig. \ref{fig:anglewidth} for reference.

 Since $b_t$ is a projection of $x_t$ onto $S_{t}$, and $m_t$ is the mid-point of $x_t,b_t$, for $u\in \cU_0$, the hyperplane $H_u'$ containing $x_t$ and parallel to $H_u$ is a supporting hyperplane for $\cC_t^-$.

Thus, using the definition of $K_u$ from \eqref{eq:projlength},
\begin{equation}\label{eq:dummy1}
\Delta_t  \le \frac{1}{V_d} \int_{\cU_0} (|\cC_{t,u}| - |\cC_{t,u}^-|) du= -\frac{||x_t-b_t||}{2V_d} \int_{\cU_0} u^T 
\frac{(x_t-m_t)}{||x_t-m_t||}  \ du,
\end{equation}
since $||x_t-m_t|| = ||x_t-b_t||/2$.

Recall the definition of $C_{w_t^\star}(c_t^\star)$ from Definition \ref{defn:anglewidth} which implies that the convex hull of $m_t$ and $S_{t}$, $\cC_t$ is contained in $C_{w_t^\star}(c_t^\star)$.
Next, we consider $\cU_1$ the set of unit vectors $u$ such that hyperplanes $H_u$ are supporting hyperplanes to $C_{w_t^\star}(c_t^\star)$ at point $m_t$ 
such that $u^T (x_t-m_t)\ge 0$. 
By definition $\cC_t\subseteq C_{w_t^\star}(c_t^\star)$, it follows that 
$\cU_1\subset \cU_0$.

Thus, from \eqref{eq:dummy1}
 \begin{equation}\label{eq:dummy2}
\Delta_t  \le -\frac{||x_t-b_t||}{2V_d} \int_{\cU_1} u^T. \frac{(x_t-m_t)}{||x_t-m_t||} du
\end{equation}

Recalling the definition of $w_t^\star$ (Definition \ref{defn:anglewidth}), 
vector $u\in \cU_1$ can be written as 
$$ u = \lambda u_{\perp} + \sqrt{1-\lambda^2} w_t^\star,$$
where $u_{\perp}^T w_t^\star=0$, $|u_{\perp}|=1$ and since $u\in \cU_1$
$$0 \le \lambda=\sqrt{1-(u^Tw_t^\star)} = u^Tu_{\perp}\le c_t^\star.$$

Let $\cS_{\perp} = \{u_\perp: |u_\perp|=1 , u_\perp^T w_t^\star=0\}$. Let $du_{\perp}$ be the 
$(n-2)$-dimensional Lebesgue measure of $\cS_{\perp}$. 

It is easy to verify that 
$du = \lambda^{d-2}(1-\lambda^2)^{-1/2} d\lambda du_{\perp}$ and hence from \eqref{eq:dummy2}

\begin{equation}\label{eq:dummy3}
\Delta_t  \le -\frac{||x_t-b_t||}{V_d} \int_{0}^{c_t^\star}  \lambda^{d-2}(1-\lambda^2)^{-1/2} d\lambda \int_{\cS_{\perp}} (\lambda u_{\perp} + \sqrt{1-\lambda^2} w_t^\star)^T \frac{(x_t-m_t)}{||x_t-m_t||}  du_{\perp}.
\end{equation}
 
 Note that $\int_{du_{\perp}} u_{\perp} du_{\perp}=0$. Thus,
 \begin{align}\nn \label{}
\Delta_t  & = -\frac{||x_t-b_t||}{2V_d} \frac{(w_t^\star)^T(x_t-m_t)}{||x_t-m_t||}  \int_{0}^{c_t^\star}  \lambda^{d-2}(1-\lambda^2)^{-1/2}   \sqrt{1-\lambda^2}\ d\lambda   \int_{\cS_{\perp}} du_{\perp},\\ 
\nn
& \stackrel{(a)}\le -V_{d-1} \frac{||x_t-b_t||}{2V_d}  \frac{ (w_t^\star)^T(x_t-m_t)}{||x_t-m_t||}  \int_{0}^{c_t^\star} \lambda^{d-2}\ d\lambda, \\ \nn
& \stackrel{(b)}\le  -V_{d-1}\frac{||x_t-b_t||}{2V_d(d-1)} c_t^\star (c_t^\star)^{d-1},\\ 
& = -V_{d-1}\frac{||x_t-b_t||}{2V_d(d-1)}  (c_t^\star)^{d},
\end{align}
where $(a)$ follows since  $\int_{\cS_{\perp}} du_{\perp} = V_{d-1}$ by definition, $(b)$ follows since $ \frac{(w_t^\star)^T(x_t-m_t)}{||x_t-m_t||} \ge c_t^\star$ from Definition \ref{defn:anglewidth}.


 \end{proof}
 

%% file: App-Switch.tex
\section{Proof of Theorem \ref{thm:main2}}
 \begin{proof}
 Since $\text{CCV}(t)$ is a monotone non-decreasing function, let $t_{\min}$ be the largest time until which Algorithm \ref{coco_alg_1} is followed by $\mathrm{Switch}$.
 The regret guarantee is easy to prove.  From Theorem \ref{thm:main1},
 regret until time $t_{\min}$ is at most $O(\sqrt{t_{\min}})$. Moreover, starting from time $t_{\min}$ till $T$, from Theorem \ref{thm:sinha2024}, the regret of Algorithm \ref{coco_sinha} is at most $O(\sqrt{T-t_{\min}})$. Thus, the overall regret for $\mathrm{Switch}$ is at most $O(\sqrt{T})$.
 
 For the CCV, with $\mathrm{Switch}$, until time $t_{\min}$, $\text{CCV}(t_{\min})\le \sqrt{T}\log T$. At 
 time $t_{\min}$, $\mathrm{Switch}$ starts to use Algorithm \ref{coco_sinha} which has the following appealing property from (8) \cite{Sinha2024} that for any $t\ge t_{\min}$ where at time $t_{\min}$  Algorithm \ref{coco_sinha} was started to be used with resetting $\text{CCV}(t_{\min})=0$. 
 For any $t\ge t_{\min}$
 \begin{eqnarray} \label{gen-fn-ineq}
		\Phi(\text{CCV}(t)) +\textrm{Regret}_t(x^\star) \leq \sqrt{\sum_{\tau=t_{\min}}^t \big(\Phi'(\text{CCV}(\tau))\big)^2} + \sqrt{t-t_{\min}}.
\end{eqnarray}
where $\beta = (2GD)^{-1}, V=1, \lambda = \frac{1}{2\sqrt{T}}, \Phi(x)= \exp(\lambda x)-1, $ and $\lambda=\frac{1}{2\sqrt{T}}$.
We trivially have $\textrm{Regret}_t(x^\star)\geq -\frac{Dt}{2D} \geq -\frac{t}{2}.$ Hence, from \eqref{gen-fn-ineq}, we have that for any $\lambda = \frac{1}{2\sqrt{T}}$ and any $t \ge t_{\min}$
$$\text{CCV}_{[t_{\min},T]} \leq 4GD\ln(2\big(1+2T)\big)\sqrt{T}.$$
Since as argued before, with $\mathrm{Switch}$,  $\text{CCV}(t_{\min})\le \sqrt{T}\log T$, we get that  $\text{CCV}_{[1:T]}\le O(\sqrt{T}\log T)$.
 \end{proof}

%% file: app-2D.tex
\section{Preliminaries for Bounding the CCV in Theorem \ref{thm:ocs} and Theorem \ref{thm:tvmonotone}}
%
%
%
%
%

Let $K_1, \dots, K_T$ be nested (i.e., $K_1 \supseteq K_2 \supset K_3 \supseteq \dots \supseteq K_T$) bounded convex subsets of $\bbR^d$. 

\begin{definition}\label{defn:projectioncurve}
If $\sigma_1\in K_1$, and $\sigma_{t+1} = \cP_{K_{t+1}}(\sigma_t)$, for $t=1, \dots, T$. Then the curve 
$${\underline \sigma}= \{(\sigma_1,\sigma_2), (\sigma_2,\sigma_3), \dots, (\sigma_{T-1},\sigma_T)\}$$ is called the projection curve on $K_1, \dots, K_T$.
\end{definition}

We are interested in upper bounding the quantity 
\begin{equation}\label{eq:totalDistance}
\Sigma = \max_{{\underline \sigma}} \sum_{t=1}^{T-1} ||\sigma_t - \sigma_{t+1}||.
\end{equation}

 \begin{lemma}\label{lem:projection}
For a projection curve ${\underline \sigma}$, $\Sigma \le d^{d/2} \text{diameter}(K_1)$.
\end{lemma}

To prove the result we need the following definition.

\begin{definition}\label{defn:se-curve} A curve $\gamma: I \rightarrow \bbR^d$  is called self-expanded, if for every $t$ where 
$\gamma'(t)$ exists, we have 
$$< \gamma'(t), \gamma(t)-\gamma(u)> \ \ge 0$$ for all $u\in I$ with $u \le t$, where $<.,.>$ represents the inner product. 
In words, what this means is that $\gamma$ starting in a point $x_0$ is self expanded, if for every $x\in \gamma$ for which there exists the tangent line $\sfT$, the arc (sub-curve) $(x_0, x)$ is
contained in one of the two half-spaces, bounded by the hyperplane through
$x$ and orthogonal to $\sfT$. 
\end{definition}
For self-expanded curves the following classical result is known.
\begin{theorem}\label{thm:manselli}\cite{Manselli}
For any self-expanded curve $\gamma$ belonging to a closed bounded convex set of $\bbR^d$ with diameter $D$, its total length is at most $O(d^{d/2} D)$.
\end{theorem}
\begin{proof}[Proof of Lemma \ref{lem:projection}]
From Definition \ref{defn:projectioncurve}, the projection curve is 
$${\underline \sigma}=\{(\sigma_1,\sigma_2), (\sigma_2,\sigma_3), \dots, (\sigma_{T-1},\sigma_T)\}.$$ Let the reverse curve be ${\underline r} = \{r_t\}_{t=0, \dots, T-2}$, where $r_t = (\sigma_{T-t}, \sigma_{T-t-1})$. Thus we are reading ${\underline \sigma}$ backwards and calling it ${\underline r}$. Note that since $\sigma_{t}$ is the projection of $\sigma_{t-1}$ on $K_t$, each piece-wise linear segment $(\sigma_t, \sigma_{t+1})$ is a straight line and hence differentiable except at the end points. Moreover, since each $\sigma_t$ is obtained by projecting $\sigma_{t-1}$ onto $K_t$ and $K_{t+1}\subseteq K_t$, we have that the projection hyperplane 
$F_t$ that passes through $\sigma_t=\cP_{K_t}(\sigma_{t-1})$ and is perpendicular to $\sigma_t - \sigma_{t-1}$ separates the two sub curves $\{(\sigma_1,\sigma_2), (\sigma_2,\sigma_3), \dots, (\sigma_{t-1},\sigma_t)\}$ and $\{(\sigma_t,\sigma_{t+1}), (\sigma_{t+1},\sigma_{t+2}), \dots, (\sigma_{T-1},\sigma_T)\}$.

Thus, we have that 
for each segment $r_\tau$, at each point where it is differentiable, the curve $r_1, \dots r_{\tau-1}$ lies on one side of the hyperplane that passes through the point and is perpendicular to $ r_{\tau+1}$. Thus, we conclude that curve ${\underline r}$ is self-expanded.


As a result, Theorem \ref{defn:se-curve} implies that the length of ${\underline r}$ is at most $O(d^{d/2} \text{diameter}(K_1))$, and the result follows since the length of ${\underline r}$ is same as that of ${\underline \sigma}$ which is $\Sigma$. 
\end{proof}

\section{Proof of Theorem \ref{thm:ocs}}
Clearly, with $f_t\equiv0$ for all $t$, with Algorithm \ref{coco_alg_1}, $y_t=x_t$ and the successive $x_t$'s are such that $x_{t+1} = \cP_{S_t}(x_{t})$. Thus, essentially, the curve ${\underline x} = (x_1, x_2), (x_2,x_3), \dots, (x_{T-1}, x_{T})$ formed by Algorithm \ref{coco_alg_1} for OCS is a projection curve (Definition \ref{defn:projectioncurve}) on $S_1\supseteq, \dots, \supseteq S_T$ and the result follows from Lemma \ref{lem:projection} and the fact that $\text{diameter}(S_1)\le D$.

\section{Proof of Theorem \ref{thm:tvmonotone}} 

%
%

\begin{proof}
Recall that $d=2$, and the definition of $F_t$ from Definition \ref{defn:projhyperplane}. Let the center be $\sfc=\cP_{S_1}(x_1)$.  Let $t_{\text{orth}}$ be the earliest $t$ for which $\angle (F_t, F_1) = \pi$.

Initialize $\kappa=1$, $s(1)=1$, $\tau(1) =1$.

{\bf BeginProcedure}
Step 1:Definition of Phase $\kappa$.
Consider $$\tau(\kappa) = \arg \max_{s(\kappa)< t \le t_{\text{orth}}, \angle(F_{s(\kappa)}, F_t) \le \pi/4} t.$$

{\bf If there is no such $\tau(\kappa)$}, 

\quad Phase $\kappa$ ends, define Phase $\kappa$ as {\bf Empty},  $s(\kappa+1) =  \tau(\kappa)+1$.

{\bf Else If} 

\quad $\angle(F_{\tau(\kappa)}, F_1)=\pi$ Exit

{\bf Else If} 

\quad $s(\kappa+1)=\tau(\kappa)$

{\bf End If}

Increment $\kappa=\kappa+1$,  and Go to Step 1.

{\bf EndProcedure}

\begin{example}\label{exm:phasedef} To better understand the definition of phases, consider Fig. \ref{fig:phases}, where the largest $t$ for which the angle between $F_t$ and $F_1$ is at most $\pi/4$ is $3$. Thus, $\tau(1)=3$, i.e., phase $1$ explores till time $t=3$ and phase $1$ ends. The starting hyperplane to consider in phase $2$ is $s(2)=3$ 
and given that angle between $F_3$ and and the next hyperplane $F_4$ is more than $\pi/4$, phase $2$ is empty and phase $2$ ends by exploring till $t=4$. The starting hyperplane to consider in phase $3$ is $s(3)=4$ and the process goes on. The first time $t$ such that the angle between $F_1$ and $F_t$ is $\pi$ is $t=6$, and thus $t_{\text{orth}}=6$, and the process stops at time $t=6$. 
This also implies that $S_6 \subset F_1$. 
Since $S_t$'s are nested, for all $t\ge 6$, $S_t\subset F_1$. Hence the total CCV after $t\ge t_{\text{orth}}$ is at most $GD$.
\end{example}

The main idea with defining phases, is to partition the whole space into empty and non-empty regions, where in each non-empty region, the starting and ending hyperplanes have an angle to at most $\pi/4$, while in an empty phase the starting and ending hyperplanes have an angle of at least $\pi/4$. Thus, we get the following simple result.

\begin{figure}

\includegraphics[width=15cm,keepaspectratio,angle=0]{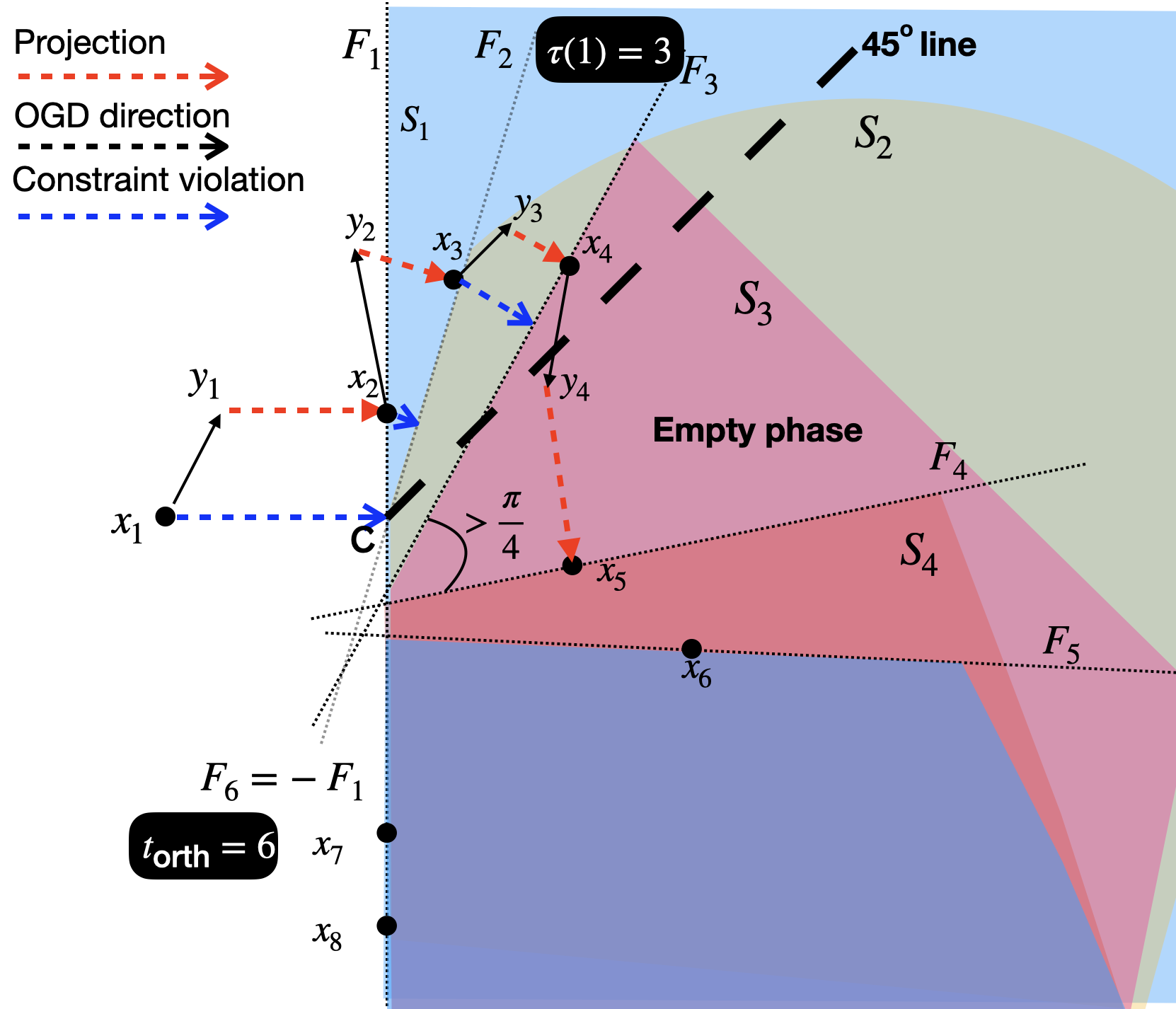}
\caption{Figure corresponding to Example \ref{exm:phasedef}.}
\label{fig:phases}
\end{figure}

\begin{lemma}\label{lem:nrphases} For $d=2$, there can be at most $4$ non-empty and $4$ empty phases.  
\end{lemma}
Proof is immediate from the definition of the phases, since any consecutively occurring non-empty and empty phase exhausts an angle of at least $\pi/4$.

\begin{rem}\label{rem:aftertorth}
Since we are in $d=2$ dimensions, for all $t\ge t_{\text{orth}}$, the movement is along the hyperplane $F_1$ and thus the resulting constraint violation after time $t\ge t_{\text{orth}}$ is at most $GD$. Thus, in the phase definition above, we have only considered time till $t_{\text{orth}}$ and we only need to upper bound the CCV till time $t_{\text{orth}}$. \end{rem}

We next define the following required quantities.

\begin{definition}\label{defn:tstar}
With respect to the quantities defined for Algorithm \ref{coco_alg_1}, let for a non-empty phase $\kappa$ 
$$r_{\max}(\kappa)= \max_{s(\kappa) < t\le \tau(\kappa)} || y_t - \sfc||\ \text{and} \ t^\star(\kappa) = \arg \max_{s(\kappa) < t\le \tau(\kappa)}^T || y_t- \sfc||.$$
%
\end{definition}
$t^\star(\kappa)$ is the time index belonging to phase $\kappa$ for which $y_t$ is the farthest.

\begin{definition} 
A non-empty phase $\kappa$ consists of time slots $\cT(\kappa) = [\tau(\kappa-1), \tau(\kappa)]$ and the angle $\angle(F_{t_1}, F_{t_2}) \le \pi/4$ for all $t_1,t_2\in \cT(\kappa)$. Using Definition \ref{defn:tstar}, we partition $\cT(\kappa)$ as $\cT(\kappa) = \cT^-(\kappa) \cup \cT^+(\kappa)$, where $\cT^-(\kappa) = [\tau(\kappa-1)+1, t^\star(\kappa)+1]$ and $\cT^+(\kappa) = [ t^\star(\kappa)+2, \tau(\kappa)]$.
\end{definition}

Thus, $\cT(\kappa)$ and $\cT(\kappa+1)$ have one common time slot.

\begin{figure}
\includegraphics[width=15cm,keepaspectratio,angle=0]{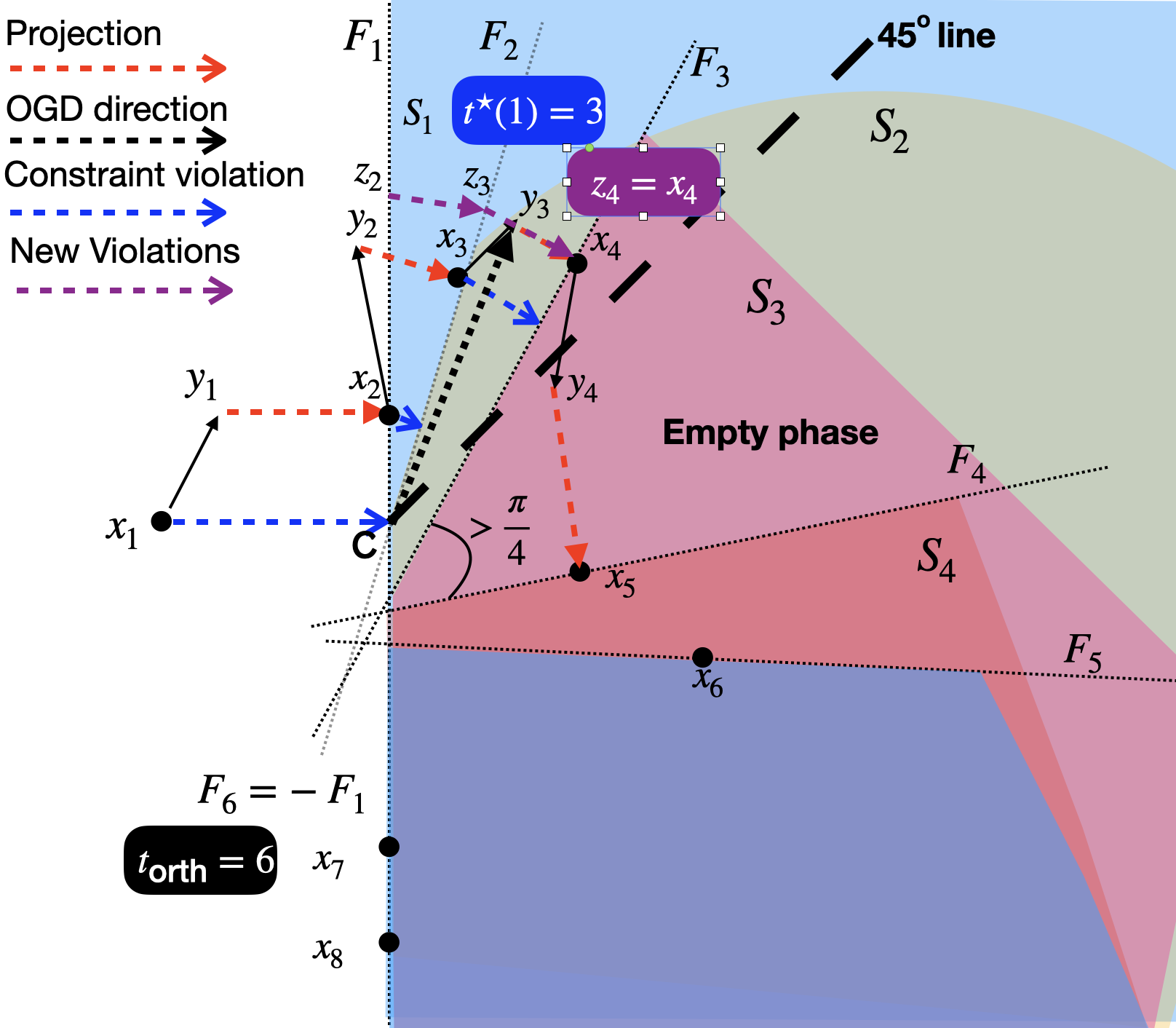}
\caption{Illustration of definition of $z_t(\kappa)$ for $t\in \cT(\kappa)$. In this example, for phase $1$, $t^\star(1)=3$ since the distance of $y_3$ from $\sfc$ is the farthest for phase $1$ that consists of time slots 
$\cT(1) = \{2,3\}$. Hence $z_{t^\star(1)+1}(1)=x_4$. For $t \in \cT(1) \backslash  t^\star(1)+1$,  $z_{t}(1)$ are such 
$z_{t+1}(1)$ is a projection of $z_{t}(1)$ onto $F_t$.}
\label{fig:MaxR}
\end{figure}
\begin{definition}\label{}
[Definition of $z_t(\kappa)$ \  for $t\in \cT^-(\kappa)$]. Let 
$z_{t^\star(\kappa)+1} = x_{t^\star(\kappa)+1}$.
For $t \in \cT^-(\kappa) \backslash t^\star(\kappa)+1$, define $z_t(\kappa)$ inductively as follows. 
$z_t(\kappa)$ is the pre-image of $z_{t+1}(\kappa)$ on $F_{t-1}$ such that the projection of $z_t(\kappa)$ on $F_t$ 
is $z_{t+1}(\kappa)$. 
\end{definition}

\begin{definition}\label{}
[Definition of $z_t(\kappa)$ \  for \ $t\in \cT^+(\kappa)$]. 
For $t\in \cT^+(\kappa)$, define $z_t(\kappa)$ inductively as follows. 
$z_t(\kappa)$ is the projection of $z_{t-1}(\kappa)$ on $F_{t-1}$. 
\end{definition}

See Fig. \ref{fig:MaxR} for a visual illustration of $t^\star(\kappa)$ and $z_t(\kappa)$.

The main idea behind defining $z_t(\kappa)$'s  is as follows. For each non-empty phase, we will construct a projection curve (Definition \ref{defn:projectioncurve}) using points $z_k$ such that the length of the projection curve upper bounds the CCV of Algorithm \ref{coco_alg_1} (shown in Lemma \ref{lem:violationub}), and then use Lemma \ref{lem:projection} to upper bound the length of the projection curve. 

\begin{definition}\label{}
[Definition of $S_t'$ for a non-empty phase $\kappa$:]  $S_{t^\star(\kappa)+1}' = S_{t^\star(\kappa)+1}$.
For $t \in \cT^-(\kappa) \backslash t^\star(\kappa)+1$, 
$S_t'$ is the convex hull of $z_{t+1}(\kappa) \cup S_t \cup S'_{t+1}(\kappa)$. For $t\in \cT^+(\kappa)$, 
$S_t' =S_t$.
See Fig. \ref{fig:defSprime}.
\end{definition}

\begin{lemma}\label{lem:nestedprime} For a non-empty phase $\kappa$, for any $t \in \cT(\kappa)$, $S_{t+1}' \subseteq S_t' $, i.e. they are nested.
\end{lemma}
\begin{figure}
\includegraphics[width=10cm,keepaspectratio,angle=0]{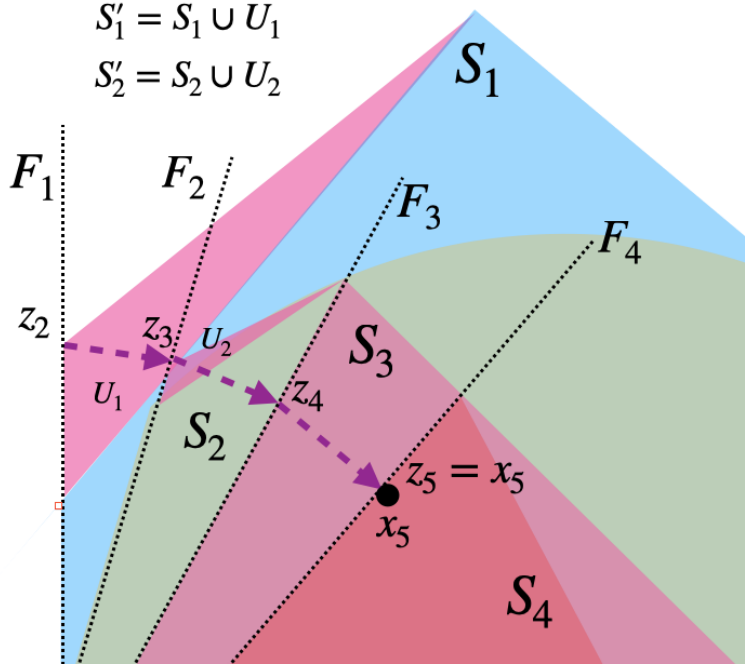}
\caption{Definition of $S_t$'s where $U_t$ are the extra regions that are added to $S_t$ to get $S_t'$.}
\label{fig:defSprime}
\end{figure}

\begin{definition} For a non-empty phase, 
 $\chi(\kappa) = S_{\tau(\kappa-1)}'  \cap \cH_{\tau(\kappa)}^+$, where $\cH_{\tau(\kappa)}^+$ has been defined in Definition \ref{defn:projhyperplane}.

\end{definition}

\begin{definition}\label{}
[New Violations for  $t\in \cT(\kappa)$:] 
For a non-empty phase $\kappa$, for $t\in \cT(\kappa) \backslash \tau(\kappa-1)$, let 
$$v_t(\kappa) = ||z_t(\kappa)-z_{t-1}(\kappa)||.$$
\end{definition}

%

\begin{lemma}\label{lem:membership} For each non-empty phase $\kappa$, all $z_{t}(\kappa)$'s for $t\in \cT(\kappa)$ belongs to $\cB(\sfc, \sqrt{2}D)$, where $\cB(c,r)$ is a ball with radius $r$ centered at $c$. In other words, $\chi(\kappa) \subseteq \cB(\sfc, \sqrt{2}D)$.
\end{lemma}
\begin{proof}
Recall that for a non-empty phase $\kappa$,  $\cT(\kappa) = \cT^-(\kappa) \cup  \cT^+(\kappa).$ We first argue about $t\in \cT^-(\kappa)$.
By definition, $z_{t^\star(\kappa)+1} = x_{t^\star(\kappa)+1}$ and $x_{t^\star(\kappa)+1}\in S_{t^\star(\kappa)}$. Thus, $z_{t^\star(\kappa)+1} \in \cB(\sfc, \sqrt{2}D)$.
Next we argue for $t \in \cT^-(\kappa) \backslash t^\star(\kappa)+1$.
Recall that the diameter of $\cX$ is $D$, and the fact that $y_t \in S_{t-1}$ from Algorithm \ref{coco_alg_1}. Thus, for any non-empty phase $\kappa$, the distance from $\sfc$ to the farthest $y_t$ belonging to the phase $\kappa$ is at most $D$, i.e., $r_{\max}(\kappa)\le D$. 
Let the pre-image of $z_{t^\star(\kappa)+1}(\kappa)$ onto $F_{s(\kappa)}$ (the base hyperplane with respect to which all hyperplanes have an angle of at most $\pi/4$ in phase $\kappa$) be $p(\kappa)$ such that projection of $p(\kappa)$ onto $F_{s(\kappa)}$ is $z_{t^\star(\kappa)+1}(\kappa)$. 
From the definition of any non-empty phase, the angle between $F_{s(\kappa)}$ and $F_{t}$ for $t\in \cT(\kappa)$ is at most $\pi/4$. 
Thus, the distance of $p(\kappa)$ from $\sfc$ is at most $\sqrt{2}D$. 


Consider the `triangle' $\Pi(\kappa)$ that is the convex hull of $\sfc, z_{t^\star(\kappa)+1}(\kappa)$ and $p(\kappa)$.
Given that the angle between $F_{t^\star(\kappa)}$ and $F_{t^\star(\kappa)-1}$ is at most $\pi/4$, the argument above implies that 
$z_t(\kappa) \in \Pi(\kappa)$ for $t=t^\star(\kappa)$. For $t= t^\star(\kappa)-1$, $z_t(\kappa) \in F_{t-1}$ is the projection of  $z_{t-1}(\kappa)$ onto $S_{t-1}'$. This implies that the distance of $z_t(\kappa)$ (for $t=t^\star(\kappa)-1$) from $\sfc$ is at most 
$$\frac{D}{\cos(\alpha_{t, t^\star(\kappa)}) \cos(\alpha_{t^\star(\kappa), t^\star(\kappa)+1})},$$ where 
$\alpha_{t_1,t_2}$ is the angle between $F_{t_1}$ and $F_{t_2}$.
From the monotonicity of angles $\theta_t$ (Definition \ref{defn:anglemonotone}), and the definition of a non-empty phase, we have that $\alpha_{t, t^\star(\kappa)}+\alpha_{t^\star(\kappa), t^\star(\kappa)+1} \le \pi/4$ and $\alpha_{t, t^\star(\kappa)}\ge 0, \alpha_{t^\star(\kappa), t^\star(\kappa)+1}\ge 0$.
Next, we appeal to the identity
\begin{equation}\label{eq:cosidentity}
\cos(A+B) \le \cos(A)\cos(B)
\end{equation} where $A+B\le \pi/4$, to claim that $z_t(\kappa) \in \Pi(\kappa)$ for $t=t^\star(\kappa)-1$. 

Iteratively using this argument while invoking the identity \eqref{eq:cosidentity} gives the result that for any $t \in \cT^-(\kappa)$, we have that $z_{t}(\kappa)$  belongs to $\Pi(\kappa)$. Since $\Pi(\kappa) \subseteq \cB(\sfc, \sqrt{2}D)$, we have the claim for all $t\in \cT^-(\kappa)$. 

%
%
%
%

By definition $z_{t}(\kappa)$ for $t\in \cT^+(\kappa)$ belong to $S_{t-1}\subseteq S_1$. Thus, their distance from $\sfc$ is at most $D$. 
\end{proof}

\begin{lemma}\label{lem:violationub} For each non-empty phase $\kappa$, and for $t\in \cT(\kappa)$ the violation $v_t(\kappa)\ge  \text{dist}(x_t, S_t)$, where $\text{dist}(x_t, S_t)$ is the original violation.
\end{lemma}
\begin{proof}
By construction of any non-empty phase $\kappa$, for $t\in \cT(\kappa)$ both $x_t(\kappa)$ and $z_t(\kappa)$ belong to $F_{t-1}$. Moreover, by construction, the distance of $z_t(\kappa)$ from $\sfc$ is at least as much as the distance of $x_t$ from $\sfc$. Thus, using  the monotonicity property of angles $\theta_t$ (Definition \ref{defn:anglemonotone}) we get the result. See Fig. \ref{fig:MaxR} for a visual illustration.
\end{proof}

For each non-empty phase $\kappa$, by definition, the curve defined by sequence $z_{t}(\kappa)$ for $t\in\cT(\kappa)$ is a projection curve (Definition \ref{defn:projectioncurve}) on sets $S'_t(\kappa)$ (note that $S'_t(\kappa)$'s  are nested from Lemma \ref{lem:nestedprime}). Moreover, for all $t\in\cT(\kappa)$, set $S'_t(\kappa) \subset \chi(\kappa)$ which is a bounded convex set. 
Thus, for $d=2$ from Lemma \ref{lem:projection} the length of curve ${\underline z}(\kappa) = \{(z_{t}(\kappa), z_{t+1}(\kappa))\}_{t\in \cT(\kappa)}$
\begin{equation}\label{eq:totallengthprojection}
\sum_{t\in \cT(\kappa)} v_t(\kappa) \le 2
\text{diameter}(\chi(\kappa)).
\end{equation}

By definition, the number of non-empty phases till time $t_{\text{orth}}$ is at most $4$. Moreover, in each non-empty phase $\chi(\kappa) \subseteq \cB(\sfc, \sqrt{2}D)$ from Lemma \ref{lem:membership} . 

Thus, from  \eqref{eq:totallengthprojection}, we have that \begin{align}\nn
\sum_{\text{Phase} \ \kappa \ \text{is non-empty}} \ \ \ \sum_{t\in \cT(\kappa)} v_t(\kappa) &\le \sum_{\text{Phase} \ \kappa \ \text{is non-empty}} 2\ \text{diameter}(\chi(\kappa)) \\ \label{eq:summeanwidth}
&\le 8 \ \text{diameter}(\cB(\sfc, \sqrt{2}D))\le O(D).
\end{align}

Using Lemma \ref{lem:violationub}, we get 
\begin{align}\label{eq:finalviolation}
\sum_{\text{Phase} \ \kappa \ \text{is non-empty}} \ \ \ \sum_{t\in \cT(\kappa)}\text{dist}(x_t, S_t) \le O(D).
\end{align}

For any empty phase, the constraint violation is the length of line segment $(x_t,\cP_{S_t}(x_{t}))$ (Algorithm \ref{coco_alg_1}) crossing it is a straight line whose length is at most $O(D)$. 
 Moreover, the total number of empty phases (Lemma \ref{lem:nrphases}) is a constant.
 Thus, the length of the curve $(x_t,\cP_{S_t}(x_{t}))$ for Algorithm \ref{coco_alg_1} corresponding to all empty phases is at $O(D)$.
%
%

Recall from \eqref{eq:distviolationrelation} that the CCV is at most $G$ times $\text{dist}(x_t, S_t)$.
Thus, from \eqref{eq:finalviolation} we get that the total violation incurred by Algorithm \ref{coco_alg_1} corresponding to non-empty phases is at most $O(GD)$, while corresponding to empty phases is at $O(GD)$.
Finally, accounting for the very first violation $\text{dist}(x_1, S_1)\le D$ and the fact that the CCV after time $t\ge t_{\text{orth}}$ (Remark \ref{rem:aftertorth}) is at most $GD$, we get that the total constraint violation $\text{CCV}_{[1:T]}$ for Algorithm \ref{coco_alg_1} is at most $O(G D)$. 

\end{proof}